\documentclass{article}

\usepackage{arxiv}

\usepackage[utf8]{inputenc} 
\usepackage[T1]{fontenc}    
\usepackage{hyperref}       
\usepackage{url}            
\usepackage{booktabs}       
\usepackage{amsfonts}       
\usepackage{nicefrac}       
\usepackage{microtype}      
\usepackage{lipsum}		
\usepackage{graphicx}
\usepackage[numbers]{natbib}
\usepackage{doi}
\usepackage{times}
\usepackage{multirow}
\usepackage{mathtools}
\newcommand{\indep}{\perp\!\!\!\!\perp}

\usepackage{amsmath}
\usepackage{amssymb}
\usepackage{natbib}
\usepackage{graphicx}
\usepackage{url}
\usepackage{algorithm2e}
\usepackage{cancel}
\usepackage{subfig}

\title{C-HDNet: Hyperdimensional Computing for Causal Effect Estimation from Observational Data Under Network Interference}

\author{
	Abhishek Dalvi 
     \And
	Neil Ashtekar 
	\And
	Vasant Honavar 
}

\renewcommand{\shorttitle}{\textit{C-HDNet}}

\hypersetup{
pdftitle={C-HDNet: Hyperdimensional Computing for Causal Effect Estimation from Observational Data Under Network Interference(preprint)},
pdfsubject={ML},
pdfauthor={Abhishek Dalvi},
pdfkeywords={Causality, Matching},
}

\begin{document}
\maketitle

\begin{abstract}
    We address the problem of estimating causal effects from observational data in the presence of network confounding, a setting where both treatment assignment and observed outcomes of individuals may be influenced by their neighbors within a network structure, resulting in network interference. Traditional causal inference methods often fail to account for these dependencies, leading to biased estimates. To tackle this challenge, we introduce a novel matching-based approach that utilizes principles from hyperdimensional computing to effectively encode and incorporate structural network information. This enables more accurate identification of comparable individuals, thereby improving the reliability of causal effect estimates. Through extensive empirical evaluation on multiple benchmark datasets, we demonstrate that our method either outperforms or performs on par with existing state-of-the-art approaches, including several recent deep learning-based models that are significantly more computationally intensive. In addition to its strong empirical performance, our method offers substantial practical advantages, achieving nearly an order-of-magnitude reduction in runtime without compromising accuracy, making it particularly well-suited for large-scale or time-sensitive applications.
\end{abstract}

\keywords{Causal Effect Estimation \and Network Data \and Network Interference \and Hyperdimensional Computing}

\section{Introduction}
Estimation of causal effects from observational data is a central concern in many disciplines, including epidemiology, economics, and social sciences, where randomized controlled experiments are often impractical or downright unethical. Hence, there is great interest in methods for drawing valid inferences about the consequences of interventions or exposures in such settings.  Unlike associations, which describe statistical relationships, causal inference aims to address counterfactual questions, e.g., would the economy have recovered in the absence of an interest rate cut? The fundamental problem of causal inference is that while we can observe the factual outcome for a given individual under one treatment assignment, the counterfactual, the outcome under any other treatment assignment, is not observed by definition \cite{Pearl_textbook, causal_neal_brady_book,hernan2010causal}.

A variety of techniques have been proposed to estimate causal effects from observational data.
The simplest of such methods estimate the counterfactual outcome for an individual of interest (the target individual) from the observed outcome(s) of other  individual(s) that are identical to the target individual in nearly every respect except the treatment status using matching techniques. \citep{stuart2010matching}. When the covariates that describe the individuals are high-dimensional, simple matching techniques break down because of the curse of dimensionality. A  variety of methods, ranging from simple dimensionality reduction to sophisticated representation learning techniques that leverage modern deep learning methods have been developed 
\citep{TARNET, Dragonnet, CFRnet, neuralmatching} (see \cite{Igelstr_CEM,yao2021survey} for reviews). 

Traditional methods for causal effect estimation from observational data typically assume independence between individuals, which is violated in settings where individuals are not independent, for example, when they are connected by social ties which introduce network interference. 
Consider for example,  a city or country that is experiencing an outbreak of a contagious disease. Social connections between individuals can be represented by a network. Suppose we want to understand how vaccination impacts the disease spread. Such understanding is crucial for creating tailored intervention strategies, such as vaccination or isolation, to  mitigate the spread of the disease within the population.  However, in the presence of social ties, the decision of an individual to get vaccinated is likely to be influenced by the decisions of his or her friends, i.e., the network of social ties  \citep{epidemic_network_example}, as a result of network interference.  

The presence of network interference presents several challenges in estimating causal effects in such settings. The most significant challenge is that an individual's outcome can be directly or indirectly influenced by the treatment status of their network neighbors. Unobserved factors that affect both individual treatment and outcomes can create spurious correlations, making it difficult to isolate the true causal effect. Networks can exhibit complex interaction patterns, including direct effects (between neighbors) and indirect effects (through multiple hops in the network).
Hence, much recent work has focused on methods for estimating causal effect from networked observational data. Almost all such methods are variants of deep learning methods for causal effect estimation from observational data that attempt to account for network interference by exploiting the knowledge of network structure  \cite{guo2020counterfactual, guo2020ignite, guo2020learning,veitch2019using}. However, deep learning techniques are computationally expensive, include many tunable hyperparameters, and require a large amount of training data to be effective \citep{Thompson_DL_Diminishing_Returns}. 

This paper aims to explore computationally efficient alternatives to deep learning methods for causal effect estimation from observational data in the presence of network interference.
Our approach leverages recent advances in hyperdimensional (HD) computing \citep{KanervaSDM, Kanerva2009_introHD}. HD computing does not require iterative optimization for training and instead relies on simple, fast, operations in a high-dimensional space. Each data sample is represented using hyperdimensional bipolar vectors in $\{-1,1\}^{\beta}$, or as multi-bit vectors, where the dimensionality is typically $\beta \approx 10,000$. Our recent work has demonstrated the effectiveness of HD computing for causal effect estimation from observational data in the simpler setting where there is no network interference and hence the data samples are independent.

In this work, we propose \emph{C-HDNet: a Causal HD computing technique for effect estimation from Network observational data}. To the best of our knowledge, this is the first application of  hyperdimensional computing for causal effect estimation from observational data in the presence of network interference. The key novelty of C-HDNet lies in the design of the hyperdimensional representation of the covariates that describe the individuals (network nodes) and the  relationships that link them to other individuals (network links) into HD vectors; and an efficient algorithm for matching the individuals based on their HD representations. The matched pairs of individuals with similar covariates that appear in similar network contexts but have differ in treatment status can then be used to estimate the individual treatment effect. The primary advantage of our approach over deep learning based  methods for causal effect estimation in the presence of network interference \cite{guo2020counterfactual, guo2020ignite, guo2020learning,veitch2019using} is that it requires no iterative optimization and operates as a one-pass learning algorithm — i.e., it involves computing specific steps without the need for repetitive training. Hence, C-HDNet offers a computationally efficient alternative to deep learning. The results of our extensive experiments demonstrate that the efficiency gains are realized without sacrificing accuracy of causal effect estimation relative to the state-of-the-art deep learning methods. 

The rest of the paper is organized as follows. Section \ref{sect:background}, provides background information on causal inference and HD computing. Section \ref{sect:problem} provides precise problem formulation. Section \ref{sect:HDNet} describes our proposed algorithm. Section \ref{sect:experiments} presents results of experiments that compare our algorithm against state-of-the-art baselines. Finally, section \ref{sect:conc} concludes the paper. 

\section{Background}\label{sect:background}

\subsection{Causal Inference Framework}
\hfill\\
The frameworks developed by \citet{rubin2005causal} and \citet{Pearl_textbook} offering distinct yet complementary foundations  for causal modeling and causal effect estimation. Rubin's framework focuses on counterfactual reasoning to estimate causal effects while Pearl's causal Bayesian networks and DAGs provide a graphical and algorithmic framework for causal inference. Both frameworks have significantly advanced our ability to infer causal relationships from observational data, each offering complementary tools and insights to address different aspects of the causal inference problem.

 In this work we follow Rubin's \textit{potential outcomes} framework. We let $T_i$ denote the treatment assignment for individual $i$. We assume treatments are binary:  $T_i$ can take values $T=1$ or $T=0$, but not both. Therefore, calculating the effect of treatment on an individual $i$ requires comparing the outcome under treatment $Y_{i}^{T=1}$ with the control outcome $Y_{i}^{T=0}$, of which only one of is observed in the data. We call the observed outcome the factual outcome, and the unobserved outcome the counterfactual outcome. 
The Individual Treatment Effect (ITE) for individual $i$ is defined as:
\[ \text{ITE}_i= Y_i^{T=1} - Y_i^{T=0}. \]
The  Average Treatment Effect (ATE) is defined as the expectation over individual treatment effects.
 \[ \text{ATE}= \mathbb{E} [Y^{T=1} - Y^{T=0}]=\mathbb {E}[Y^{T=1}]- \mathbb {E}[Y^{T=0}].\] 

Randomized Control Trials (RCTs) are the gold standard for estimating causal effects. Random assignment ensures that all variables, known and unknown, are equally distributed between the groups in expectation. This setup simplifies the calculation of the Average Treatment Effect (ATE). Unlike observational studies where differences in outcomes between treated and untreated groups might be confounded by other factors, in an RCT, the ATE is essentially the expected difference between the observed outcomes of the treatment and control groups. Therefore, in RCT, estimating the causal effect of treatment reduces to assessing the association between treatment and outcome.

However, RCT are not always feasible due to their high cost or ethical concerns. 
Hence, there is much interest in methods for reliably estimating causal effects from observational data. The following assumptions suffice for estimating causal effects from observational data \citep{balacningrsenbaum, no_interference,SUTVA_cite}: 

\begin{enumerate}
\item {\em Ignorability}, which implies that the treatment assignment $T$ is independent of the potential outcomes $Y^{T=0}$ and $Y^{T=1}$ given the covariates ${\bf X}$. This means that $(Y^{T=0}, Y^{T=1}) \indep T|{\bf X}={\bf x}$.
\item {\em Positivity/Overlap}, indicating that the propensity scores should be between 0 and 1, i.e., $0 < P(T = 1|{\bf X}={\bf x}) < 1$ for all values ${\bf x}$ of the covariates ${\bf X}$.
\item {\em Non-interference}, $Y_{i}(T_{1}, \cdots, T_{i}) = Y_{i}(T_i)$, implying that the outcome of any individual is unaffected by the treatment of all other individuals.
\item {\em Consistency}, for all $t$, if $T_i = t$, then $Y_i(t) = Y_i$, ensuring that the observed effect of the assigned treatment is equal to the potential outcome of that treatment and is necessary for estimating causal effects from observational data.
\end{enumerate}
Under these conditions, it can be shown that: \[ \text{ATE}= \mathbb {E}_{\bf X} [\mathbb {E}[Y|T=1, {\bf X}={\bf x}]- \mathbb {E}[Y |T=0,{\bf X}={\bf x}]]. \] 

To estimate causal effects, one must account for potential confounding due to covariates through \textit{covariate adjustment}. One of the simplest techniques for adjusting covariates is \textit{matching}, where the goal is to find a nearly identical ``twin'' for each individual with the opposite treatment status. Numerous matching techniques have been proposed, based on nearest neighbors, propensity scores \citep{Rubin_matchedpropensity, king2019propensity} and balancing scores \citep{balacningrsenbaum}. A comprehensive overview of matching techniques can be found in \citet{stuart2010matching}.

Recently, there has been growing interest in deep neural networks for covariate adjustment due to their predictive power. Techniques such as CFRnet \citep{CFRnet} learn latent representations for the covariates, minimizing the distributional differences between control and treatment populations. Dragonnet \citep{Dragonnet} leverages Rubin's lemma \citep{Rubin_matchedpropensity} and makes use of propensity scores with targeted regularization, while borrowing concepts from \citet{TARNET} and \citet{AIPW_double}. These methods aim to mitigate confounding bias within the original covariates when under binary treatments.

Overall, the preceding covariate adjustment techniques rely on estimating the outcome using conditional outcome modeling estimators, i.e., $\hat{\mu}(\mathbf{x}, 1) \approx \mathbb{E}[Y \mid T=1, \mathbf{X}=\mathbf{x}]$ and $\hat{\mu}(\mathbf{x}, 0) \approx \mathbb{E}[Y \mid T=0, \mathbf{X}=\mathbf{x}]$. The estimator $\hat{\mu}$ can range from a simple model such as Nearest Neighbors regressor to a more complex model such as a deep neural network.

\subsection{Hyperdimensional Computing}\hfill

\noindent Hyperdimensional Computing \citep{KanervaSDM, Kanerva2009_introHD, Neubert2019_introHD}, offers an attractive alternative to deep learning methods. It represents features as extremely high-dimensional binary or bipolar vectors -- often around 10,000 dimensions -- on which simple element wise operations are performed for tasks including learning and inference. Unlike deep learning, which relies on iterative minimization of a loss function, HD computing follows predefined steps to map data in a single pass to  to a high-dimensional space.

In this work, we employ either transformed/hashed or randomly generated bipolar hyperdimensional vectors, denoted as $\{-1,1\}^{\beta}$ with $\beta = 10,000$. A collection of such vectors is subsequently used and transformed through hyperdimensional operations to produce multi-bit representations used for downstream inference. To measure similarity between hypervectors, we use a distance metric $d_{\mathcal{H}}$: for bipolar vectors, this is typically the Hamming distance; for multi-bit hypervectors, the L2 distance is used. In both cases, cosine similarity may also serve as an alternative similarity measure.

A key property of randomly generated bipolar hypervectors is their near-orthogonality: the similarity between any two such vectors, $\mathbf{a}$ and $\mathbf{b}$, is $d_{\mathcal{H}}(\mathbf{a}, \mathbf{b}) \approx 0.5$ with high probability. This orthogonality, along with the way HD computing distributes information across all dimensions through vector operations, makes hyperdimensional computing an appealing framework for machine learning \citep{KanervaSDM}. Its inherent robustness to noise and support for content-based retrieval offer advantages over traditional deep learning approaches \citep{Neubert2019_introHD}. In our study, we make use of the following HD computing operations:

\begin{itemize}

    \item \textbf{Binding} $\otimes$: is dimension-wise multiplication operation between HD vectors. The resulting vector is dissimilar from the original inputs. For example, given two randomly sampled bipolar vectors $\mathbf{a}$ and $\mathbf{b}$, if we compute $\mathbf{c} = \mathbf{a} \otimes \mathbf{b}$, then the distances between the result and the original vectors satisfy $d_{\mathcal{H}}(\mathbf{c}, \mathbf{a}) \approx 0.5$ and $d_{\mathcal{H}}(\mathbf{c}, \mathbf{b}) \approx 0.5$, indicating dissimilarity/orthogonal nature for the Binding operation.\\
    
    \item \textbf{Bundling} \( \oplus \): is a dimension-wise operation between HD vectors that performs addition across corresponding dimensions. Bundling ensures that the resulting vector, derived from sampled HD vectors within a set, is similar to all the sampled vectors while being dissimilar to non-sampled HD vectors. Further, Bundling is both associative and commutative, meaning: $\mathbf{a} \oplus (\mathbf{b} \oplus \mathbf{c}) = (\mathbf{a} \oplus \mathbf{b}) \oplus \mathbf{c} = (\mathbf{c} \oplus \mathbf{a}) \oplus \mathbf{b}.$ Bundling is often used as an aggregate representation of a set of hyper vectors. In some literature, Bundling is performed as signed addition, while in this work we use dimension-wise addition, following the approach used by \citet{RelHD}.

\end{itemize}

\noindent Combining Bundling and Binding enables concise representations of data with complex features. For instance, consider a dataset containing information about people. Consider one individual ``Alice'', who is of age 25, height of 170, and weight of 120. Alice's features can be represented using an information-preserving hypervector as follows: $
Alice_{HV} = (Age_{HV} \otimes 25_{HV}) \oplus (Height_{HV} \otimes 170_{HV}) \oplus (Weight_{HV} \otimes 120_{HV})
$. The other individuals in the dataset can be encoded in a similar manner. For a more in-depth overview please refer to \citet{Kanerva2009_introHD} and \citet{Neubert2019_introHD}.

\section{Problem Statement}\label{sect:problem}

We are given a dataset with $N$ individuals, with each individual $i \in \{1, \dots, N\}$ represented by a $d$-dimensional feature vector $\mathbf{x}_i \in \mathbb{R}^{d}$. We assume treatments are binary:  $T_i \in \{ 0,1 \}$ denotes the observed treatment, and $Y^{(T_i)}_i \in \mathbb{R}$ denotes the corresponding observed (factual) outcome for individual $i$. 
We use $\mathbf{A}$ to denote the adjacency matrix of the network, which represents connections between individuals. We assume that the network is undirected and that all edges are unweighted, i.e., $ A_{ij} = A_{ji}$ and $\mathbf{A} \in \{ 0,1 \}^{N\times N}$.

\begin{figure}[t]
\includegraphics[width=4cm]{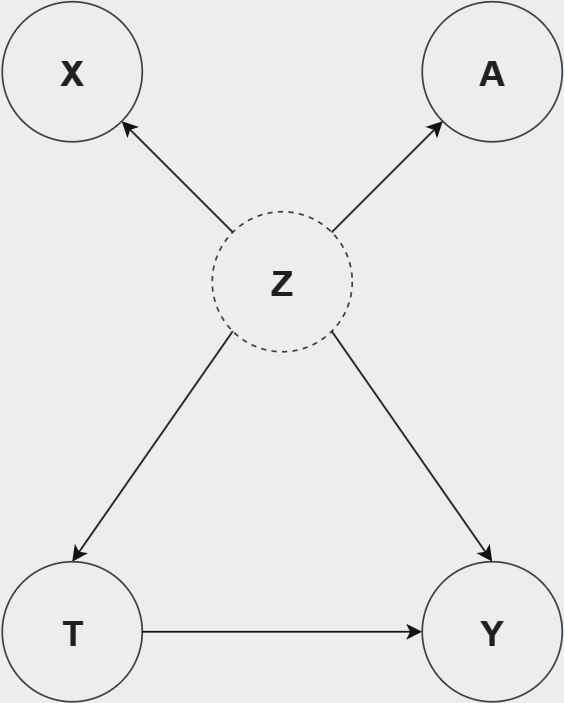}
\centering
\caption{The causal diagram for the networked observational data problem setting. The covariates or features for nodes/individuals are represented by $\mathbf{X}$, the network structure is represented by $\mathbf{A}$. 
$\mathbf{Z}$ is a latent representation, thus represented by a dotted circle. Ideally, conditioning on $\mathbf{Z}$ should deconfound the causal effect of treatment $\mathbf{T}$ on outcome $\mathbf{Y}$. }
\label{causal_digram_matching_spatio}
\end{figure}

In this work, we assume that the treatment assignment and outcomes for individual $i$ are  causally influenced by $\mathcal{N}^{k}(i)$, which represents the $k$-hop neighbors of node $i$. In this setting, some of the assumptions listed in Section \ref{sect:background} are violated:
\begin{itemize}
   \item \emph{Violation of the Non-interference Assumption}: 
   $Y_i(T_1, ..., T_i, ..., T_{N}) \not= Y_i(T_i)$
  \item \emph{Violation of the Conditional Ignoribility Assumption}: $ Y_i(0), Y_i(1) \not\indep T_i|\mathbf{x}_{i}$
\end{itemize}

\noindent Given that we assume an individual's treatment assignment and outcomes are causally influenced by $\mathbf{x}_{i}$ and $\mathcal{N}^{k}(i)$, we posit the existence of a latent variable $\mathbf{Z}$ acting as a confounder. This latent variable serves as a proxy for both the individual covariates $\mathbf{X}$ and the network structure $\mathbf{A}$. Conditioning on $\mathbf{Z}$ mitigates the confounding effects of treatment $\mathbf{T}$ on outcome $\mathbf{Y}$.

\noindent Therefore, given observational data $\mathcal{D} = \left( \left\lbrace (\mathbf{x}_i, T_i, Y^{(T_i)}_i) \right\rbrace_{i=1}^N, \mathbf{A} \right)$, our objective is to represent the latent variable $\mathbf{Z}=f(\mathbf{X},\mathbf{A})$ as function of individual features $\mathbf{X}$ and network connections $\mathbf{A}$. This ensures that the condition $Y^{(1)}_{i}, Y^{(0)}_{i} \indep T_{i} \mid \mathbf{z}_{i}$ or $Y^{(1)}_{i}, Y^{(0)}_{i} \indep T_{i} \mid (\mathbf{x}_{i}, \,\, \mathcal{N}^{k}(i))$ is satisfied $\forall i \in \{1 \cdots N\}$, enabling unbiased estimation of individual treatment effects. Refer to Fig. \ref{causal_digram_matching_spatio} for a causal diagram illustrating this problem setting.

\section{C-HDNet}\label{sect:HDNet}

Conventional covariate adjustment techniques -- which condition only on $\mathbf{X}$ -- are inadequate for our problem setting as they fail to address confounding from $\mathbf{Z}$. Techniques proposed by \citet{guo2020learning, guo2020counterfactual, guo2020ignite} address this issue by modeling the latent variable $\mathbf{Z}$, which serves as a proxy for both $\mathbf{X}$ and $\mathbf{A}$, using Graph Neural Networks \citep{kipf2016semi, veličković2018graph, hamilton2017inductive}. These networks are used because of their strong predictive performance across a wide range of applications \citep{chen2020graph, wu2020comprehensive}.

Our primary goal is to design a model that is both low-latency and performant. To achieve this goal, we first map the original covariates to a latent representation which incorporates network information, then we perform nearest neighbor matching. 

\subsection{Mapping Covariates to Hyperdimensional Representations}
\hfill\\
To make use of HD computing, we must first represent our data with hyperdimensional vectors. To do so, we start by mapping our covariates $\mathbf{x}_{i} \in \mathbb{R}^{d}$ using Random Hyperplane Tessellations \citep{Vershyninhyperplane_tessla, Dirksensharp_randomHyperPlane,dalvi_RHPT_causal}. Specifically, we map our covariates to hyperdimensional vectors $\mathbf{r}_{i}$ using the following equation:
\begin{equation}\label{RHPT}
    \mathbf{r}_{i} = \text{sign}(\mathbf{Q}\mathbf{x}_i + \Gamma).
\end{equation}

\noindent Here, $\Gamma$ is sampled from a uniform distribution over $[-\lambda, \lambda]^{\beta}$, and $\mathbf{Q}$ is a matrix in $\mathbb{R}^{\beta \times d}$ with rows drawn from a $d$-dimensional normal distribution.

Previous research by \citet{dalvi_RHPT_causal} demonstrated that if 
$Y^{(1)}_{i}, Y^{(0)}_{i} \indep T_{i} \mid \mathbf{x}_i$, then the transformation in Eq. \ref{RHPT} serves as a valid approximate balancing score \citep{balacningrsenbaum}. Specifically, \citet{dalvi_RHPT_causal} showed that $\forall i \in \{1 \cdots N\}$, $(Y^{(1)}_{i}, Y^{(0)}_{i}) \indep T_{i} \mid \mathbf{x}_i \implies (Y^{(1)}_{i}, Y^{(0)}_{i}) \indep T_{i} \mid \mathbf{r}_i$ approximately holds with high probability when $\beta$ is sufficiently large and the true propensity score function $e(\mathbf{x}_{i}) = \mathbb{E}[T=1|\mathbf{x}_{i}]$ is smooth. This conclusion is based on the findings that the mapping from $\mathbf{x}_{i}$ to $\mathbf{r}_{i}$ (i) preserves distances with extremely high probability when $\beta$ is sufficiently large and (ii) provides no additional information regarding treatment assignment, thereby maintaining conditional independence.

In our setting, the network itself may act as a confounder. This could result in a scenario where $Y_i(1)$ and $Y_i(0)$ are not independent of $T_i$ given $\mathbf{r_i}$. Regardless, since we assume that $Y^{(1)}_{i}, Y^{(0)}_{i} \indep T \mid (\mathbf{x}_{i}, \mathcal{N}^{k}(i))$; we can assert that $Y^{(1)}, Y^{(0)} \indep T \mid (\mathbf{r}_{i}, \mathcal{N}^{k}(i))$. Again, this is because the mapping in Eq. \ref{RHPT} preserves ignorability.

To enable matching to deconfound the effect of the network structure, we  design a function that utilizes HD computing. This function incorporates network information using hypervectors, resulting in a latent representation which can then be used for matching.

\subsection{Incorporating Network Structure}\hfill

\noindent We now describe how network structure is incorporated in our HD representation. We make use of the RelHD method proposed in \citet{RelHD}. RelHD constructs a representation $\mathbf{z}_i$ for node $i$ as follows:

 \begin{equation} \label{latent-eq}
    \mathbf{z}_{i} = (\psi_{0} \otimes \mathbf{r}_{i}) \oplus (\psi_{1} \otimes \mathbf{h}^{\text{1-hop}}_{i}) \oplus (\psi_{2} \otimes \mathbf{h}^{\text{2-hop}}_{i})
\end{equation}

\noindent where $\psi_{0}$, $\psi_{1}$, and $\psi_{2}$ are randomly-sampled bipolar hypervectors with fixed values across all nodes $i \in \{1, \dotsc, N\}$. The $\mathbf{h}_i$ terms are constructed as:

$$\mathbf{h}^{\text{1-hop}}_{i} = \mathop{\oplus}_{j \in \mathcal{N}^{1}(i)} \mathbf{r}_{j}$$

$$\mathbf{h}^{\text{2-hop}}_{i} = \mathop{\oplus}_{j \in \mathcal{N}^{1}(i)} \mathbf{h}^{\text{1-hop}}_{j}$$

\noindent In this encoding, $\mathbf{h}^{\text{1-hop}}_{i}$ aggregates the 1-hop neighborhood information of node $i$, while $\mathbf{h}^{\text{2-hop}}_{i}$ aggregates the 2-hop neighborhood information of node $i$ by propagating the previously-aggregated 1-hop information. The $\mathbf{r}_i$, $\mathbf{h}^{\text{1-hop}}_{i}$, and $\mathbf{h}^{\text{2-hop}}_{i}$ vectors are binded to the $\psi_{0}$, $\psi_{1}$, and $\psi_{2}$ vectors in order to indicate relative network position. The resulting terms are bundled, aggregating node-level and neighborhood-level information. Note that this encoding method inherently accommodates sparse networks and disconnected nodes. For example, if there exist a node $p$ which is disconnected, the latent representation of node $p$ will be $\mathbf{z}_{p} = (\psi_{0} \otimes \mathbf{r}_{p})$, since $\mathbf{h}^{\text{1-hop}}_{p} = \mathbf{h}^{\text{2-hop}}_{p} = [0]^{10000}$, since $\mathcal{N}^{1}(p) = \emptyset$

Our inclusion of 1-hop and 2-hop neighbor information is motivated by the literature on graph neural networks (GNNs). Including information from 1-hop and 2-hop neighbors is typically sufficient for performing downstream classification or regression tasks while avoiding the GNN \textit{oversmoothing} problem \citep{oversmoothing-gnn}. Oversmoothing occurs when samples with meaningfully different labels are mapped to similar latent representations. Incorporating information from distant nodes (beyond 2-hop) can result in oversmoothing, which may limit predictive performance. 

We hypothesize that our constructed representation $\mathbf{z}_{i}$ possesses sufficient information to approximately satisfy conditional ignorability, i.e., $Y^{(1)}, Y^{(0)} \indep T\vert \mathbf{Z}$, given the causal model in Fig. \ref{causal_digram_matching_spatio}. Assuming this condition holds, matching can be performed to faithfully estimate treatment effects. 

The primary advantage of our approach over existing methods is that ours requires \textit{no training} and operates as a one-pass learning algorithm. Unlike state-of-the-art deep learning techniques, our approach does not involve iterative optimization (i.e. no backpropagation or gradient descent), resulting in far greater computational efficiency. 

\subsection{Nearest Neighbor Matching for Outcome Prediction}
\hfill

\noindent Matching, as discussed in \citet{stuart2010matching}, is a method used to mitigate bias from confounding variables when estimating causal effects in observational data. In our approach, we perform matching using the latent representation \(\mathbf{Z}\) in order to account for the covariates as well as the network. This method effectively reduces bias from observed confounders and operates efficiently through weighted $k$-nearest neighbors regression. Note that we use $m$ rather than $k$ to denote the number of neighbors and avoid abuse of notation\footnote{Recall that $\mathcal{N}^{k}(i)$ represents the $k$-hop neighbors of node $i$ in the remainder of the paper.}.

To calculate the counterfactual outcome for an individual \(i\), we use m-NN\(_{1}\) and m-NN\(_{0}\), which represent the $m$-nearest neighbors from the treatment and control groups, respectively. The outcomes are computed as follows:

\begin{equation*}
\begin{split}
    \widehat{Y}^{1}_{i} &= \sum_{j \in \text{m-NN}_{1}(\mathbf{z}_{i})} w_{j} Y_{j}; \\
    \widehat{Y}^{0}_{i} &= \sum_{j \in \text{m-NN}_{0}(\mathbf{z}_{i})} w_{j} Y_{j}
\end{split}
\end{equation*}

\noindent Here, $w_{j}$ denotes weights assigned based on distance (i.e., nearer neighbors are assigned more weight) under the constraint that $\sum_{j}w_{j} = 1$.

\section{ Experiments and Results}\label{sect:experiments}
We proceed to describe the data sets used, the experimental setup, the state-of-the art baselines used in our experiments, and results of comparison of C-HDNet with state-of-the-art methods for causal effect estimation from observational data in the presence of network interference. We also analyze the effect of $k$, the number of hops used to incorporate information from network neighbors of each node into its HD representation. 

The code base for our experiments can be accessed using the following link: \newline\url{https://github.com/Abhishek-Dalvi410/C-HDNet}.

\subsection{Data Sets}\hfill

\begin{table}[h!]
\centering
\caption{Summary of BlogCatalog and Flickr graph datasets.}
\begin{tabular}{lcc}
\toprule
\textbf{Statistic} & \textbf{BlogCatalog} & \textbf{Flickr} \\
\midrule
Number of Nodes & 5,196 & 7,575 \\
Dimensionality of Features & 2,198 & 1,205 \\
Number of Links & 171,743 & 239,738 \\
Average Node Degree & 66.11 & 63.30 \\
Median Node Degree & 49 & 25 \\
Node Degree (10th Percentile) & 22 & 7 \\
Node Degree (90th Percentile) & 134 & 147\\
\bottomrule
\end{tabular}
\label{tab:dataset-stats}
\end{table}

\noindent As observational data does not include counterfactual outcomes, synthetic or semi-synthetic data is typically used in the evaluation of causal effect estimation techniques. We perform evaluations on two datasets -- BlogCatalog and Flickr -- used in \citet{guo2020counterfactual, guo2020learning}. An overview of the data sets is provided in Table~\ref{tab:dataset-stats}; the spread of the node degrees — captured by the average, median, and percentile values —highlight the realistic nature of these data sets, showcasing sparsely and densely connected regions.

\begin{itemize}
    \item \textbf{BlogCatalog}: In this dataset, each node represents a blogger, with links indicating social relationships between bloggers. The node features includes processed bag-of-words representations derived from keywords extracted from blogger descriptions. The treatment or intervention pertains to whether a blogger's content receives more views on mobile devices ($T=1$) or desktops ($T=0$). The outcomes indicate the opinions of readers for each blogger. 

    \item \textbf{Flickr}: This dataset represents an online social network where users share images and videos. Nodes are users and each edge signifies a social connection (friendship) between a pair of users. Users' features are constructed from a list of interest tags on images and videos. Following the same methodology and assumptions as applied to the BlogCatalog dataset,  user's content is viewed on mobile devices ($T=1$) or desktops ($T=0$). The outcomes indicate the opinions of readers on each user profile. 
\end{itemize}

\noindent We adopt the data generation process  described in \citet{guo2020learning, guo2020counterfactual} for the setting where a user's features and those of their neighbors influence treatment assignment and reader opinions (see Figure \ref{causal_digram_matching_spatio}). We want to know how receiving more views on mobile devices, compared to desktops, impacts reader opinions -- an exploration of individual treatment effects in this context. We evaluate the three versions of each dataset provided by \citet{guo2020learning, guo2020counterfactual} with the confounding factor for the individual (node) covariates set to $\kappa_0=10$, while the confounding factor for the covariates of the 1-hop neighbors are set to $\kappa_1 = \{0.5, 1, 2\}$. The higher the value of the confounding factor, the stronger the influence of the confounding of neighboring nodes on the overall effect.

Additionally, we include in our analyses confounding effects from 2-hop neighbors, denoted as $\kappa_2$ (refer to Eq. (11) in \citet{guo2020learning}, where an additional term for 2-hop neighbors is included). To better reflect real-world scenarios, we randomize the confounding factors for each node -- $\{ \kappa^{i}_0, \kappa^{i}_1, \kappa^{i}_2\} $ sampling them a Uniform distribution $\mathcal{U}; \forall i \in \{1 \cdots N\}$. In contrast, \citet{guo2020learning, guo2020counterfactual} considered the confounding factors to have same values across all nodes. For the Flickr dataset, we sample $\kappa^i_1 \sim \mathcal{U}(0.5, 3)$, while for the BlogCatalog data set, $\kappa^i_1 \sim \mathcal{U}(0.5, 1)$. In both data sets $\kappa^i_0 \sim \mathcal{U}(5, 10), \kappa^i_2 \sim \mathcal{U}(0.01, 0.05)$.

\subsection{Evaluation Metrics}\hfill

\noindent We evaluate C-HDNet against other existing methods using two primary categories of performance metrics: (1) the error in estimated causal effects compared to their true values, and (2) the time required for training and inference.

For the assessment of causal effect estimation error; we use the error on the Average Treatment Effect denoted by $\epsilon_{\text{ATE}} $ and the Rooted Precision in Estimation of Heterogeneous Effect (PEHE) error \citep{Hill_BART}  denoted by $\epsilon_{\text{PEHE}}$. These errors are estimated as follows:

$$\epsilon_{\text{ATE}} = |(\frac{1}{N} \sum_{i=1}^{N}\widehat{\text{ITE}}_{i} )- (\frac{1}{N} \sum_{i=1}^{N}\text{ITE}_{i})|$$

$$
\epsilon_{\text{PEHE}} = \sqrt{\frac{1}{N}\sum_{i=1}^{N}(\widehat{\text{ITE}}_{i}- \text{ITE}_{i})^{2}}
$$

\noindent where  and $\text{ITE}_{i} = Y^{1}_{i} - Y^{0}_{i}$ denotes the true Individual Treatment Effect and  $\widehat{\text{ITE}}_{i} = \widehat{Y}^{1}_{i} - \widehat{Y}^{0}_{i}$ denotes the Estimated Individual Treatment Effect each for the $i^{th}$ sample.

Furthermore, we conduct our experiments in two contexts: (1) \emph{In-sample}, where we estimate the causal effect using the available factual result, and (2) \emph{Out-of-sample}, where our goal is to estimate the causal effect in the absence of both potential outcomes. For all of our experiments, we split our data so that 20\% of the samples are used for assessing the effectiveness of the method beyond the sample used for training the model(s), and  the remaining 80\% of the samples are used for within-sample evaluation. 

\subsection{Baseline Methods}

We compare C-HDNet against the following baselines:

\begin{itemize}
    \item TARNet \citep{TARNET, CFRnet}: Treatment-Agnostic Representation Networks learns representations of confounders by transforming the original features into a latent space using a neural network. This network is then split into two parts to predict potential outcomes. TARNet is trained to minimize the error in inferred factual outcomes while also working to reduce the discrepancy across confounder representations over the treated and controlled groups. In this work, we use the maximum mean discrepancy (MMD) variant for balancing penalties. 
    
    \item DRGNet \citep{Dragonnet}: DragonNet similar to TARNet in terms of its network architecture. However, the network is trained to to minimize the error in inferred factual outcomes while forcing latent representations to be suggestive of the propensity score using targeted regularization, based on non-parametric estimation theory.
    
    \item C-VAE \citep{louizos2017causal_VAE}: Causal Effect Variational Autoencoder is a neural network latent that employs latent variable modeling to concurrently estimate the hidden latent space that summarizes the confounders as well as the causal effect. This method is based on Variational Autoencoders (VAE) \cite{kingma2013auto}, which adhere to the causal structure of inference using proxies.
    
    \item C-Forest \citep{wager2018estimation_Cforest}: Causal Forest is a non-parametric method for estimating heterogeneous treatment effects which extends the widely use Random forest algorithms to incorporate causal effect estimation.
    
    \item BART \citep{Hill_BART, Chipman_BART}: Bayesian Additive Regression Trees are a Bayesian nonparametric modeling procedure that allows for flexible modeling and accommodates complex relationships between confounders, treatment, and outcomes to effectively estimate causal effects.
    
    \item NetDeconf \citep{guo2020counterfactual}: Network Deconfounder shares a similar architecture to TARNet, and is trained using the same objective function. However, it utilizes graph neural networks to map covariates into latent representations, which incorporate network information in the latent space to assist in network deconfounding effects.
\end{itemize}

\begin{table}[h!]
\centering
\caption{In-sample mean errors and standard errors across 10 simulations of the BlogCatalog dataset generated by \citet{guo2020learning, guo2020counterfactual}.}
\begin{tabular}{ |c||c|c||c|c||c|c| }
\hline
  & \multicolumn{2}{|c||}{$\kappa_1 = 0.5$} & \multicolumn{2}{|c||}{$\kappa_1 = 1$} & \multicolumn{2}{|c|}{$\kappa_1 = 2$} \\
\hline
Method & $\epsilon_{\text{ATE}}$ & $\epsilon_{\text{PEHE}}$ & $\epsilon_{\text{ATE}}$ & $\epsilon_{\text{PEHE}}$ & $\epsilon_{\text{ATE}}$ & $\epsilon_{\text{PEHE}}$ \\
\hline\hline
TARNet & 1.4 $\pm$ 0.4 & 8.4 $\pm$ 0.6 & 2.0 $\pm$ 0.3 & 14.2 $\pm$ 0.6 & 4.4 $\pm$ 0.9 & 28.6 $\pm$ 0.6 \\
DRGNet & 1.4 $\pm$ 0.3 & 6.8 $\pm$ 0.3 & 1.8 $\pm$ 0.3 & 11.8 $\pm$ 0.2 & 3.9 $\pm$ 0.9 & 24.0 $\pm$ 0.3 \\
C-VAE & 1.8 $\pm$ 0.2 & 8.5 $\pm$ 0.1 & 5.7 $\pm$ 0.5 & 16.3 $\pm$ 0.4 & 9.4 $\pm$ 1.7 & 32.2 $\pm$ 1.2 \\
C-Forest & 4.4 $\pm$ 1.2 & 8.3 $\pm$ 1.5 & 3.5 $\pm$ 0.7 & 8.5 $\pm$ 1.0 & 11.9 $\pm$ 2.2 & 22.3 $\pm$ 2.6 \\
BART & 5.9 $\pm$ 1.4 & 10.4 $\pm$ 2.1 & 6.4 $\pm$ 0.9 & 10.9 $\pm$ 0.9 & 17.4 $\pm$ 3.2 & 28.2 $\pm$ 3.0 \\
NetDeconf & 0.7 $\pm$ 0.2 & 4.7 $\pm$ 0.8 & 1.3 $\pm$ 0.2 & 4.7 $\pm$ 0.5 & 2.1 $\pm$ 0.5 & 9.1 $\pm$ 1.2 \\
C-HDNet & 0.6 $\pm$ 0.1 & 2.7 $\pm$ 0.1 & 0.8 $\pm$ 0.2 & 4.1 $\pm$ 0.2 & 2.0 $\pm$ 0.4 & 7.4 $\pm$ 0.3 \\
\hline
\end{tabular}
\label{BlogCatalog-insample}
\end{table}

\begin{table}[h!]
\centering
\caption{Out-of-sample mean errors and standard errors across 10 simulations of the BlogCatalog dataset generated by \citet{guo2020learning, guo2020counterfactual}.}
\begin{tabular}{ |c||c|c||c|c||c|c| }
\hline
  & \multicolumn{2}{|c||}{$\kappa_1 = 0.5$} & \multicolumn{2}{|c||}{$\kappa_1 = 1$} & \multicolumn{2}{|c|}{$\kappa_1 = 2$} \\
\hline
Method & $\epsilon_{\text{ATE}}$ & $\epsilon_{\text{PEHE}}$ & $\epsilon_{\text{ATE}}$ & $\epsilon_{\text{PEHE}}$ & $\epsilon_{\text{ATE}}$ & $\epsilon_{\text{PEHE}}$ \\
\hline\hline
TARNet & 1.7 $\pm$ 0.5 & 10.8 $\pm$ 1.4 & 2.5 $\pm$ 0.4 & 16.4 $\pm$ 1.3 & 4.9 $\pm$ 1.2 & 35.3 $\pm$ 2.3 \\
DRGNet & 1.3 $\pm$ 0.3 & 7.4 $\pm$ 1.2 & 1.8 $\pm$ 0.3 & 9.7 $\pm$ 1.0 & 3.8 $\pm$ 1.0 & 23.0 $\pm$ 2.5 \\
C-VAE & 6.0 $\pm$ 0.9 & 11.7 $\pm$ 1.5 & 8.9 $\pm$ 1.2 & 14.4 $\pm$ 1.3 & 21.3 $\pm$ 2.0 & 30.6 $\pm$ 2.6 \\
C-Forest & 4.5 $\pm$ 1.3 & 8.3 $\pm$ 1.5 & 3.5 $\pm$ 0.7 & 8.6 $\pm$ 1.0 & 11.6 $\pm$ 2.3 & 22.9 $\pm$ 2.9 \\
BART & 1.5 $\pm$ 0.4 & 3.9 $\pm$ 0.5 & 2.4 $\pm$ 0.5 & 5.5 $\pm$ 0.5 & 8.3 $\pm$ 2.9 & 12.6 $\pm$ 2.8 \\
NetDeconf & 0.7 $\pm$ 0.2 & 4.7 $\pm$ 0.8 & 1.3 $\pm$ 0.2 & 4.7 $\pm$ 0.5 & 2.1 $\pm$ 0.4 & 9.1 $\pm$ 1.3 \\
C-HDNet & 0.3 $\pm$ 0.1 & 4.4 $\pm$ 0.7 & 0.7 $\pm$ 0.1 & 4.8 $\pm$ 0.4 & 1.2 $\pm$ 0.3 & 8.8 $\pm$ 0.9 \\
\hline
\end{tabular}
\label{BlogCatalog-outsample}
\end{table}

\subsection{Experimental Setup}

\noindent All experiments were conducted on Google Colab Pro using the high RAM version, with all models utilizing only CPU and no GPUs for computation. For C-HDNet, we chose $\beta = 10,000$ dimensions for our hyperdimensional representations, as this number has been shown to be sufficient for the properties of HD computing to hold \citep{Kanerva2009_introHD, Neubert2019_introHD, dalvi_RHPT_causal}. We used the KNN regressor from scikit-learn for matching, keeping all parameters at their default settings including the number of nearest neighbors $m=5$. Furthermore, we set $\Gamma = 0$ in Eq. \eqref{RHPT}. According to the findings of \citet{Dirksensharp_randomHyperPlane}, the parameter $\Gamma$ is necessary to distinguish or assign different binary representations to data points that lie along the same ray or concentrated along a line. However, our datasets happen to be well-distributed and scattered, making this distinction unnecessary.

\begin{table}[h!]
\centering
\caption{In-sample mean errors and standard errors across 10 simulations of the Flickr dataset generated by \citet{guo2020learning, guo2020counterfactual}.}
\begin{tabular}{ |c||c|c||c|c||c|c| }
\hline
  & \multicolumn{2}{|c||}{$\kappa_1 = 0.5$} & \multicolumn{2}{|c||}{$\kappa_1 = 1$} & \multicolumn{2}{|c|}{$\kappa_1 = 2$} \\
\hline
Method & $\epsilon_{\text{ATE}}$ & $\epsilon_{\text{PEHE}}$ & $\epsilon_{\text{ATE}}$ & $\epsilon_{\text{PEHE}}$ & $\epsilon_{\text{ATE}}$ & $\epsilon_{\text{PEHE}}$ \\
\hline\hline
TARNet & 2.5 $\pm$ 0.6 & 21.5 $\pm$ 1.5 & 4.4 $\pm$ 1.1 & 40.7 $\pm$ 1.9 & 8.8 $\pm$ 1.2 & 78.6 $\pm$ 4.0 \\
DRGNet & 1.6 $\pm$ 0.5 & 11.1 $\pm$ 0.3 & 3.1 $\pm$ 0.6 & 21.7 $\pm$ 0.6 & 5.7 $\pm$ 1.7 & 43.8 $\pm$ 1.2 \\
C-VAE & 0.6 $\pm$ 0.1 & 13.0 $\pm$ 0.0 & 2.6 $\pm$ 0.4 & 24.8 $\pm$ 0.5 & 4.1 $\pm$ 1.1 & 46.4 $\pm$ 1.0 \\
C-Forest & 3.1 $\pm$ 0.7 & 8.6 $\pm$ 0.4 & 7.9 $\pm$ 1.3 & 16.4 $\pm$ 1.0 & 14.2 $\pm$ 2.5 & 31.0 $\pm$ 1.5 \\
BART & 3.9 $\pm$ 0.4 & 9.8 $\pm$ 0.3 & 5.1 $\pm$ 0.8 & 16.8 $\pm$ 0.8 & 8.2 $\pm$ 1.4 & 31.4 $\pm$ 1.2 \\
NetDeconf & 0.7 $\pm$ 0.1 & 4.2 $\pm$ 0.2 & 1.2 $\pm$ 0.1 & 6.0 $\pm$ 0.2 & 2.3 $\pm$ 0.2 & 9.7 $\pm$ 0.4 \\
C-HDNet & 0.7 $\pm$ 0.0 & 3.1 $\pm$ 0.0 & 1.0 $\pm$ 0.0 & 6.0 $\pm$ 0.4 & 1.4 $\pm$ 0.1 & 11.9 $\pm$ 1.6 \\
\hline
\end{tabular}
\label{Flickr-insample}
\end{table}

\begin{table}[h!]
\centering
\caption{Out-of-sample mean errors and standard errors across 10 simulations of the Flickr dataset generated by \citet{guo2020learning, guo2020counterfactual}.}
\begin{tabular}{ |c||c|c||c|c||c|c| }
\hline
  & \multicolumn{2}{|c||}{$\kappa_1 = 0.5$} & \multicolumn{2}{|c||}{$\kappa_1 = 1$} & \multicolumn{2}{|c|}{$\kappa_1 = 2$} \\
\hline
Method & $\epsilon_{\text{ATE}}$ & $\epsilon_{\text{PEHE}}$ & $\epsilon_{\text{ATE}}$ & $\epsilon_{\text{PEHE}}$ & $\epsilon_{\text{ATE}}$ & $\epsilon_{\text{PEHE}}$ \\
\hline\hline
TARNet & 3.8 $\pm$ 0.9 & 28.4 $\pm$ 2.1 & 6.5 $\pm$ 1.7 & 54.7 $\pm$ 3.8 & 13.2 $\pm$ 2.0 & 99.4 $\pm$ 5.8 \\
DRGNet & 2.2 $\pm$ 0.7 & 10.2 $\pm$ 0.6 & 3.4 $\pm$ 0.7 & 18.3 $\pm$ 1.7 & 8.5 $\pm$ 3.1 & 36.2 $\pm$ 2.5 \\
C-VAE & 1.0 $\pm$ 0.1 & 14.3 $\pm$ 0.4 & 2.8 $\pm$ 0.4 & 24.8 $\pm$ 1.1 & 8.3 $\pm$ 0.6 & 40.4 $\pm$ 1.4 \\
C-Forest & 2.9 $\pm$ 0.7 & 8.8 $\pm$ 0.4 & 7.8 $\pm$ 1.3 & 16.8 $\pm$ 1.2 & 14.3 $\pm$ 2.5 & 32.1 $\pm$ 1.9 \\
BART & 2.2 $\pm$ 0.5 & 4.8 $\pm$ 0.3 & 2.5 $\pm$ 0.9 & 6.2 $\pm$ 0.6 & 5.3 $\pm$ 0.9 & 9.9 $\pm$ 0.8 \\
NetDeconf & 0.7 $\pm$ 0.1 & 4.2 $\pm$ 0.3 & 1.2 $\pm$ 0.2 & 6.1 $\pm$ 0.3 & 2.5 $\pm$ 0.2 & 9.8 $\pm$ 0.4 \\
C-HDNet & 0.5 $\pm$ 0.0 & 4.6 $\pm$ 0.3 & 0.7 $\pm$ 0.0 & 6.9 $\pm$ 0.5 & 0.9 $\pm$ 0.1 & 13.0 $\pm$ 2.0 \\
\hline
\end{tabular}
\label{Flickr-outsample}
\end{table}

For TARNet and Dragonnet, the architecture consists two layers for the latent representation followed by two layers in each of the two heads for outcome predictions. The hidden units and layers are structured as [512, 256, [128, 128], [128, 128]] for TARNet and [200, 200, [100, 100], [100, 100]] for DragonNet. All loss terms in both models are weighted equally. For CEVAE, we utilized the implementation from the causalml library \citep{chen2020causalml}, using the hyperparameters specified in the example implementation on the documentation page. In the case of Causal Forest, we set the hyperparameters to include 20 trees, with a split ratio of 0.7, a minimum of 5 observations of each type (treated and untreated) allowed in a leaf, and a maximum tree depth of 20. For BART and NetDeconf, we used the implementations provided by \citet{guo2020learning}.

\begin{table}[h!]
\centering

\caption{Mean errors and standard errors across 10 simulations of Flickr and BlogCatalog datasets for random levels of 0-hop, 1-hop, and 2-hop confounding.}
\begin{tabular}{ |c|c||c|c|c|c| }
\hline

 & & \multicolumn{2}{|c|}{In-sample} & \multicolumn{2}{|c||}{Out-of-sample} \\
 \hline

 & & $\epsilon_{\text{ATE}}$ & $\epsilon_{\text{PEHE}}$  & $\epsilon_{\text{ATE}}$  & $\epsilon_{\text{PEHE}}$  \\
\hline

\multirow{3}{*}{BlogCat} & NetDeconf & 0.3 $\pm$ 0.0 & 1.9 $\pm$ 0.0 & 0.3 $\pm$ 0.0 & 1.0 $\pm$ 0.0 \\

 & C-HDNet & 0.2 $\pm$ 0.0  & 1.9 $\pm$ 0.0 & 0.2 $\pm$ 0.0 & 1.4 $\pm$ 0.0 \\

 \hline

\multirow{3}{*}{Flickr}  & NetDeconf & 0.2 $\pm$ 0.0  & 4.0 $\pm$ 0.0 & 0.3 $\pm$ 0.0 & 3.0 $\pm$ 0.1 \\

 & C-HDNet &  0.3 $\pm$ 0.0 & 4.2 $\pm$ 0.1 & 0.2 $\pm$ 0.0 & 3.8 $\pm$ 0.1 \\
\hline

\end{tabular}

\label{Random_confound-table}
\end{table}

\subsection{Comparison with the State-of-the-Art}\hfill

\noindent The results under predefined levels of confounding for the BlogCatalog dataset are presented in Tables~\ref{BlogCatalog-insample} and~\ref{BlogCatalog-outsample}, corresponding to in-sample and out-of-sample performance, respectively. Similarly, the in-sample and out-of-sample results for the Flickr dataset are shown in Tables~\ref{Flickr-insample} and~\ref{Flickr-outsample}. On each simulation, our proposed method (C-HDNet) either outperforms all existing methods or performs similarly to the best-performing existing method. As expected, the methods which do not incorporate network information (TARNet, DRGNet, C-VAE, C-Forest, and BART) generally perform worse than the methods which do incorporate network information (NetDeconf, C-HDNet). Further, the error in effect estimation generally increases as the amount of network confounding ($\kappa_1$, confounding from 1-hop neighbors) increases.

\begin{figure}[h!]
    \centering
    \begin{minipage}{0.48\textwidth}
        \centering
        \includegraphics[width=\linewidth]{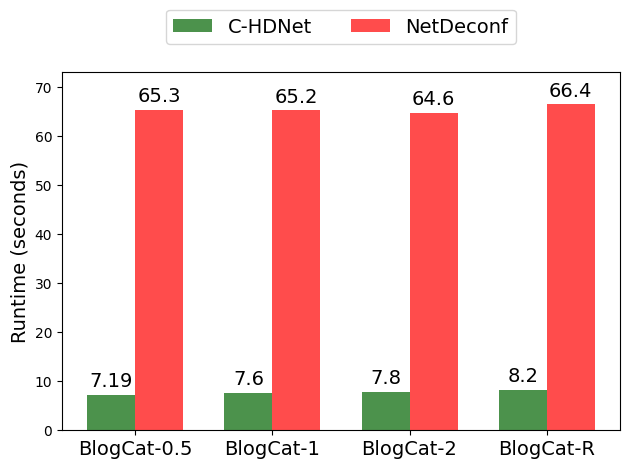}
        \textbf{(a)} BlogCatalog
    \end{minipage}
    \hfill
    \begin{minipage}{0.48\textwidth}
        \centering
        \includegraphics[width=\linewidth]{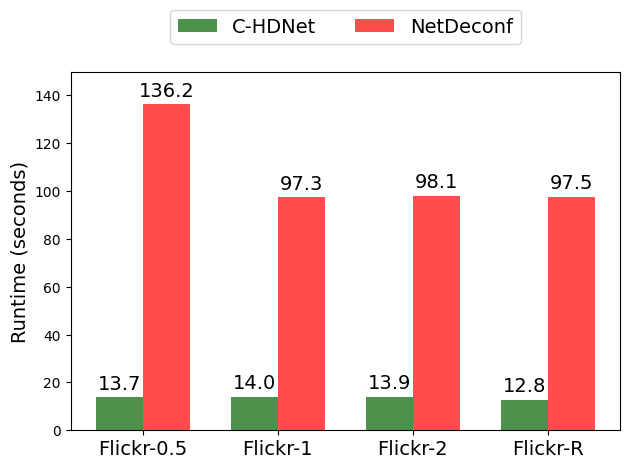}
        \textbf{(b)} Flickr
    \end{minipage}

    \caption{Runtime comparison between C-HDNet and NetDeconf across different datasets. BlogCat-R and Flickr-R refer to the datasets generated using randomly sampled 0-hop, 1-hop, and 2-hop confounding factors, while the remaining datasets are from \citet{guo2020learning, guo2020counterfactual}.}
    \label{runtime-compare}
\end{figure}

Our results for randomized confounding are given in Table \ref{Random_confound-table}. Here, we compare the best performing existing method (NetDeconf) against our proposed method (C-HDNet). We observe similar performance across methods for both the BlogCatalog and Flickr datasets. 

As a final comparison against existing methods, we consider computational efficiency (combined runtime for training and inference) rather than predictive performance. Again, we compare C-HDNet against NetDeconf, as both methods incorporate network information and offer similar predictive performance. Results are illustrated in Figure \ref{runtime-compare}. C-HDNet is faster than NetDeconf by almost an order of magnitude on the BlogCatalog simulations. On the Flickr simulations,C-HDNet also has a lower runtime than NetDeconf by a large margin. These results can be explained by the fact that C-HDNet is a simple algorithm based on dimension-wise operations and matching, while NetDeconf is a comparatively more complex algorithm requiring iterative optimization.

\subsection{Evaluating C-HDNet with varying k-Hop configurations}

\begin{figure}[h!]
    \centering
    \begin{minipage}{0.9\textwidth}
        \centering
        \includegraphics[width=\linewidth]{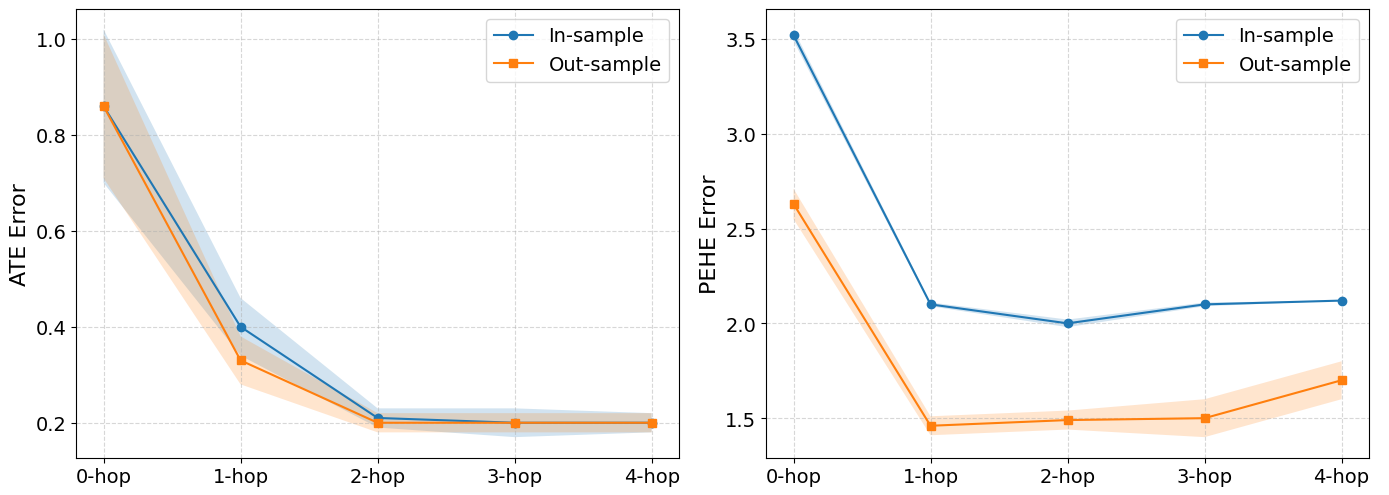}
        
        \textbf{(a)} BlogCatalog
    \end{minipage}
    
    \vspace{0.5cm}  
    
    \begin{minipage}{0.9\textwidth}
        \centering
        \includegraphics[width=\linewidth]{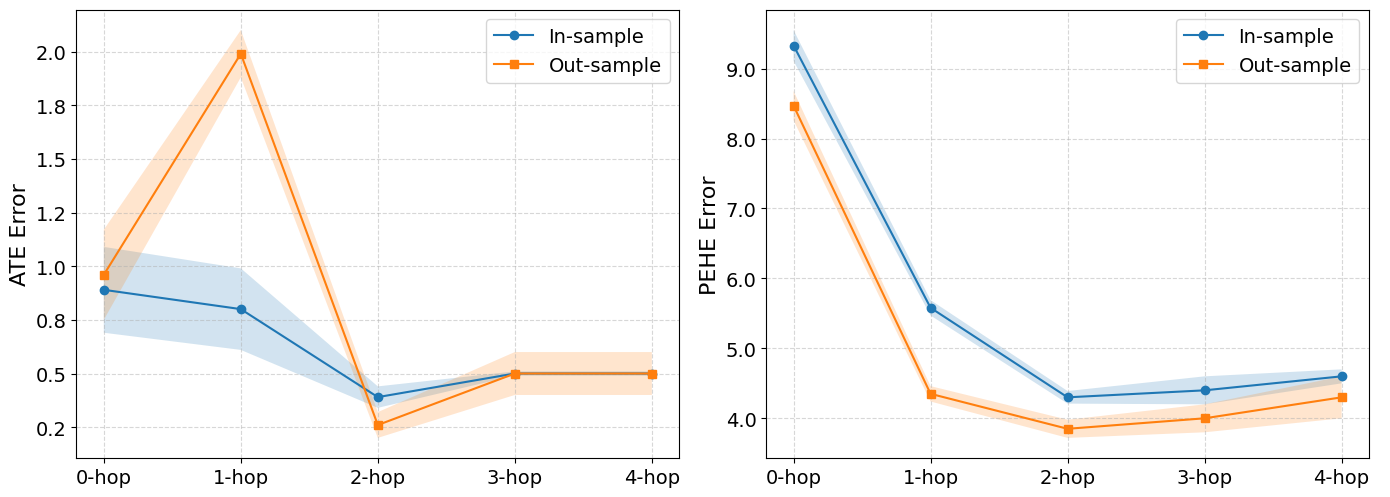}
        
        \textbf{(b)} Flickr
    \end{minipage}
    
    \caption{Mean ATE and PEHE errors of C-HDNet with increasing $k$-hop neighborhood information. Shaded areas indicate the standard error across 10 simulations.}
    \label{ablation-plot}
\end{figure}

\noindent In this section, we design experiments to evaluate how incorporating network information influences the performance of our C-HDNet algorithm. We consider HD representations of covariates under the following configurations: (1) without any network information (0-hop), (2) using up to 1-hop neighbor information, (3) using up to 2-hop neighbor information (i.e., the proposed model), (4) using up to 3-hop neighbor information, and (5) using up to 4-hop neighbor information.

In the 0-hop setting, we set $\mathbf{z}_i = (\psi_{0} \otimes \mathbf{r}_i)$, meaning only the node’s own covariates are used. For the 1-hop configuration, we remove the $\psi_2$ and $\mathbf{h}^{2\text{-hop}}_i$ terms from Eq.\eqref{latent-eq}. In contrast, for the 3-hop and 4-hop cases, we extend Eq.\eqref{latent-eq} by introducing the additional terms $(\psi_3 \otimes \mathbf{h}^{3\text{-hop}}_i)$ and $(\psi_4 \otimes \mathbf{h}^{4\text{-hop}}_i)$, respectively. The 2-hop setting corresponds to our original proposed model, where $\mathbf{z}_i$ is defined as in Eq.~\eqref{latent-eq}.

Results are summarized in Figure \ref{ablation-plot}. Random values of confounding $\{\kappa_0, \kappa_1, \kappa_2\}$ are used for the Flickr and BlogCatalog datasets -- this setup is the same as the one used to generate results for Table \ref{Random_confound-table}. 

We observe that the predictive performance of C-HDNet is generally strongest when using the full model defined in Eq.~\eqref{latent-eq}, which includes up to 2-hop neighborhood information. Performance is weakest when no network information (0-hop) is included. Notably, ATE error on the Flickr dataset deviates from this trend—the worst performance is observed when only 1-hop neighborhood information is used.

Including additional information from 3-hop and 4-hop neighborhoods also results in a performance decline compared to the 2-hop model. This degradation is likely due to the introduction of additional confounding factors or oversmoothing, as discussed in Section~\ref{sect:HDNet}. However, the decline is not as severe as that seen in the 0-hop and 1-hop settings.

In summary, omitting key neighborhood information (as in the 0-hop and 1-hop cases) leads to a substantial drop in performance due to missing confounding signals. On the other hand, incorporating excessive neighborhood depth (3-hop and 4-hop) may introduce noise or irrelevant dependencies, thereby harming predictive accuracy.

\section{Summary and Discussion}
\label{sect:conc}

In this work, we have introduced C-HDNet -- an algorithm for causal effect estimation given networked observational data. Network data is ubiquitous in the real-world, and incorporating this information when performing causal inference can help avoid confounding and improve causal effect estimates. Design of the C-HDNet algorithm was motivated by ideas from hyperdimensional computing and graph neural networks. C-HDNet first constructs a latent representation which incorporates network information, then performs matching in the latent space. The results of extensive  experiments show that our algorithm either matches or outperforms the state-of-the-art approaches at a fraction of their computational cost. 

Looking ahead, several promising directions remain for further research. One important extension is adapting C-HDNet to handle temporal or longitudinal network data, where each node in the network has observations over time, potentially with irregular intervals. This would enable the model to estimate dynamic causal effects in evolving networks.

Another limitation of the current work has to do with the lack of uncertainty quantification. At present, we provide only point estimates of causal effects without offering confidence intervals or other measures of uncertainty. Standard errors derived from repeated simulations on synthetic datasets serve as  proxy measures of the uncertainty of our causal effect estimates , but more robust techniques are needed. Estimating uncertainty is particularly challenging in networked data due to the violation of the i.i.d. assumption—interconnected nodes make traditional resampling or bootstrapping methods inappropriate. Hence, developing methods for uncertainty quantification in this context remains an important direction for future research.

\section*{Acknowledgements}

This work was funded in part by grants from the National Science Foundation (2226025),
the National Center for Advancing Translational Sciences, and the National Institutes of Health
(UL1 TR002014).

\bibliographystyle{plainnat}
\bibliography{references}  

\end{document}